%% file: main.tex
\documentclass{article}

\PassOptionsToPackage{numbers, compress}{natbib}

\usepackage[preprint]{neurips_2022}

\usepackage[utf8]{inputenc} 
\usepackage[T1]{fontenc}    
\usepackage{hyperref}       
\usepackage{url}            
\usepackage{booktabs}       
\usepackage{amsfonts}       
\usepackage{nicefrac}       
\usepackage{microtype}      
\usepackage{xcolor}         
\usepackage{algorithm}
\usepackage{algorithmic}
\usepackage[pdftex]{graphicx}     
\usepackage{mathtools}
\usepackage{svg}

\usepackage{caption}
\usepackage{subcaption}

\newcommand\tcd[0]{TRACT}

\title{\tcd: Denoising Diffusion Models with \\ Transitive Closure Time-Distillation}

\author{%
  David ~Berthelot\thanks{correspondance to dberthelot@apple.com}\\
  Apple\\
  \And Arnaud Autef\\
  Apple\\
  \And Jierui Lin\\
  Apple\\
  \And Dian Ang Yap\\
  Apple\\
  \And Shuangfei Zhai\\
  Apple\\
  \And Siyuan Hu\\
  Apple\\
  \And Daniel Zheng\\
  Apple\\
  \And Walter Talbott\\
  Apple\\
  \And Eric Gu \\
  Apple\\
}

\begin{document}

\maketitle

\begin{abstract}
Denoising Diffusion models have demonstrated their proficiency for generative sampling.
However, generating good samples often requires many iterations.
Consequently, techniques such as binary time-distillation (BTD) have been proposed to reduce the number of network calls for a fixed architecture.
In this paper, we introduce TRAnsitive Closure Time-distillation (\tcd{}), a new method that extends BTD.
For single step diffusion, \tcd{} improves FID by up to 2.4$\times$ on the same architecture, and achieves new single-step Denoising Diffusion Implicit Models (DDIM) state-of-the-art FID (7.4 for ImageNet64, 3.8 for CIFAR10).
Finally we tease apart the method through extended ablations.
The PyTorch \citep{pytorch} implementation will be released soon.
\end{abstract}

\input{introduction.tex}
\input{related_work.tex}
\input{method.tex}

\input{experiments.tex}
\input{conclusion.tex}

\subsection*{Acknowledgements}
We would like to thank Josh Susskind, Xiaoying Pang, Miguel Angel Bautista Martin and Russ Webb for their feedback and suggestions.

\subsection*{Contributions}
Here are the authors contributions to the work: 
David Berthelot led the research and came up with the transitive closure method and working code prototypes.
Arnaud Autef obtained CIFAR-10 results, designed and ran ablation experiments, set up multi-gpu and multi-node training via DDP.
Walter Talbott helped with ablation experiments and with writing.
Daniel Zheng worked on cloud compute infrastructure, set up multi-gpu and multi-node training via DDP, and ran experiments.
Siyuan Hu implemented the FID, integrated the BTD paper's model into transitive closure framework and conducted the experiments of distillation to smaller architectures.
Jierui Lin finalized data, training and evaluation pipeline, obtained 64x64 ImageNet results, integrated BTD's teacher models and noise schedule to our pipeline, reproduced binary distillation and its variants for ablation.
Dian Ang Yap implemented EDM variants, and designed experiments for TRACT (VE-EDM) on CIFAR-10 and ImageNet. 
Shuangfei Zhai contributed to the discussions, writing and ablation studies. 
Eric Gu contributed to writing and conducted experiments for distillation to smaller architectures.

\bibliographystyle{plainnat}
\bibliography{ref}


\newpage
\appendix
\input{appendix.tex}
\end{document}

%% file: introduction.tex
\section{Introduction}
 Diffusion models~\citep{sohl2015deep,song2019generative,ddpm} represent state-of-the-art generative models for many domains and applications. They work by learning to estimate the score of a given data distribution, which in practice can be implemented with a denoising autoencoder following a noise schedule. Training a diffusion model is arguably much simpler compared to many alternative generative modeling approaches, e.g., GANs~\citep{gan}, normalizing flows~\cite{dinh2017density} and auto-regressive models~\cite{imagegpt}. The loss is well-defined and stable; there is a large degree of flexibility to design the architecture; and it directly works with continuous inputs without the need for discretization. These properties make diffusion models demonstrate excellent scalability to large models and datasets, as shown in recent works in diverse domains such as image generation \cite{ho2022cascaded, dhariwal2021diffusion}, image or audio super-resolution \cite{lee2021nu, han2022nu, li2022srdiff, saharia2022image}, audio and music synthesis \cite{liu2021diffsinger, mittal2021symbolic, kong2020diffwave, chen2020wavegrad, okamoto2021noise}, language models \cite{li2022diffusion, gong2022diffuseq, hoogeboom2021argmax, austin2021structured}, and cross-domain applications such as text-to-image and text-to-speech \cite{ramesh2022hierarchical, imagen, jeong2021diff, popov2021grad,ldm}

Despite the empirical success, inference efficiency remains a major challenge for diffusion models.
As shown in~\citep{song_sde}, the inference process of diffusion models can be cast as solving a neural ODE~\citep{neural_ode}, where the sampling quality improves as the discretization error decreases.
As a result, up to thousands of denoising steps are used in practice in order to achieve high sampling quality.
This dependency on a large number of inference steps makes diffusion models less favorable compared to one-shot sampling methods, e.g., GANs, especially in resource-constrained deployment settings. 

Existing efforts for speeding up inference of diffusion models can be categorized into three classes: (1) reducing the dimensionality of inputs~\citep{ldm,vahdat2021score,gu2023fdm}; (2) improving the ODE solver~\citep{karras2022elucidating,lu2022dpm}; and (3) progressively distilling the output of a teacher diffusion model to a student model with fewer steps~\citep{binarydistill,meng2022distillation}.
Among these, the progressive distillation approach is of special interest to us.
It uses the fact that with a Denoising Diffusion Implicit Model (DDIM) inference schedule~\citep{ddim}, there is a deterministic mapping between the initial noise and the final generated result. This allows one to learn an efficient student model that approximates a given teacher model.
A naive implementation of such distillation would be prohibitive, as for each student update, the teacher network needs to be called $T$ times (where $T$ is typically large) for each student network update. \citet{binarydistill} bypass this issue by performing progressive binary time distillation (BTD).
In BTD, the distillation is divided into $\log_2(T)$ phases, and in each phase, the student model learns the inference result of two consecutive teacher model inference steps.
Experimentally, BTD can reduce the inference steps to four with minor performance loss on CIFAR10 and 64x64 ImageNet.

In this paper, we aim to push the inference efficiency of diffusion models to the extreme: one-step inference with high quality samples.
We first identify critical drawbacks of BTD that prevent it from achieving this goal: 1) objective degeneracy, where the approximation error accumulates from one distillation phase to the next, and 2) the prevention of using aggressive stochastic weights averaging (SWA) \citep{swa} to achieve good generalization, due to the fact that the training course is divided into $\log_2(T)$ distinct phases.

Motivated by these observations, we propose a novel diffusion model distillation scheme named TRAnsitive Closure Time-Distillation (\tcd).
In a nutshell, \tcd{} trains a student model to distill the output of a teacher model's inference output from step $t$ to $t'$ with $t' < t$.
The training target is computed by performing one step inference update of the teacher model to get $t \to t - 1$, followed by calling the student model to get $t - 1 \to t'$, in a bootstrapping fashion.
At the end of distillation, one can perform one-step inference with the student model by setting $t=T$ and $t'=0$.
We show that \tcd{} can be trained with only one or two phases, which avoids BTD's objective degeneracy and incompatibility with SWA. 

Experimentally, we verify that \tcd{} drastically improves upon the state-of-the-art results with one and two steps of inference.
Notably, it achieves single-step FID scores of 7.4 and 3.8 for 64x64 ImageNet and CIFAR10 respectively. 

%% file: related_work.tex
\section{Related Work}

\paragraph{Background}
DDIMs~\citep{ddim} are a subclass of Denoising Diffusion Probabilistic Models (DDPM)~\citep{ddpm} where the original noise is reused at every step $t$ of the inference process.
Typically DDIMs use a $T$-steps noise schedule $\gamma_t\in[0,1)$ for $t\in\{1,\ldots,T\}$.
By convention, $t=0$ denotes the noise-free step and therefore $\gamma_0=1$.
In the variance preserving (VP) noisification setting, a noisy sample $x_t$ is produced from the original sample $x_0$ and some Gaussian noise $\epsilon$ as follows:
\begin{equation}
x_t = x_0\sqrt{\gamma_t}+\epsilon\sqrt{1-\gamma_t}\label{eq:xt}
\end{equation}
A neural network $f_\theta$ is trained to predict either the signal, the noise or both.
The estimations of $x_0$ and $\epsilon$ at step $t$ are denoted as $x_{0|t}$ and $\epsilon_{|t}$. 
For the sake of conciseness, we only detail the signal prediction case.
During the denoisification phase, the predicted $x_{0|t}$ is used to estimate $\epsilon_{|t}$ by substitution in Equation (\ref{eq:xt}):
$$x_{0|t} \coloneqq f_\theta(x_t,t)\textrm{ and }\epsilon_{|t}=\frac{x_t - x_{0|t}\sqrt{\gamma_t}}{\sqrt{1-\gamma_t}}$$
These estimates allow inference, by substitution in Equation (\ref{eq:xt}), of $x_{t'}$ for any $t'\in\{0,\ldots,T\}$:
\begin{equation}
x_{t'}=\delta(f_\theta,x_t,t,t')\coloneqq x_t \frac{\sqrt{1-\gamma_{t'}}}{\sqrt{1-\gamma_t}} + f_\theta(x_t,t)\frac{\sqrt{\gamma_{t'}(1-\gamma_t)}-\sqrt{\gamma_t(1-\gamma_{t'})}}{\sqrt{1-\gamma_t}}\label{eq:step-function}
\end{equation}
Here we introduced the step function $\delta(f_\theta,x_t,t,t')$ to denote DDIM inference from $x_t$ to $x_{t'}$. 
\paragraph{Advanced ODE solvers} A common framework in the  denoisification process is to use stochastic differential equations (SDEs) that maintain the desired distribution $p$ as the sample $x$ evolves over time \cite{karras2022elucidating, song_sde}. Song et. al. presented a corresponding probability flow ordinary differential equation (ODE) with the initial generated noise as the only source of stochasticity. Compared to SDEs, ODEs can be solved with larger step sizes as there is no randomness between steps.

Another advantage of solving probability flow ODEs is that we can use existing numerical ODE solvers to accelerate sampling in the denoisification phase.
However, solving ODEs numerically approximates the true solution trajectory due to the truncation error from the solver. Popular numerical ODE solvers include first-order Euler's method and higher-order methods such as Runge-Kutta (RK) \cite{suli2003introduction}.
Karras et. al. apply Heun's $2^{\text{nd}}$ order method \cite{ascher1998computer} in the family of explicit second-order RK to maintain a tradeoff between truncation error and number of function evaluations (NFEs) \cite{karras2022elucidating, jolicoeur2021gotta, dormand1980family}.

However, existing ODE solvers are unable to generate high-quality samples in the few-step sampling regime (we loosely define few-steps regime in $\approx 5$ steps).
RK methods may suffer from numerical issues with large step sizes \cite{hochbruck2010exponential, hochbruck2005explicit}.
Our work provides an orthogonal direction to these ODE solvers, and \tcd{} outputs can be further refined with higher-order methods.


\paragraph{Diffusion model distillation}
The idea of distilling a pretrained diffusion model to a single step student is first introduced in ~\citep{luhman2021knowledge}.
Despite encouraging results, it suffers from high training costs and sampling quality degradation.
This idea is later extended in ~\citep{binarydistill,huang2022prodiff,meng2022distillation}, where one progressively distills a teacher model to a student by reducing its total steps by a factor of two. 

Specifically, in Binary Time-Distillation (BTD) \cite{binarydistill}, a student network $g_\phi$ is trained to replace two denoising steps of the teacher $f_\theta$.
Using the step function notation, $g_\phi$ is modeled to hold this equality:
\begin{equation}
\delta(g_\phi,x_t,t,t-2)\approx x_{t-2}\coloneqq\delta(f_\theta,\delta(f_\theta,x_t,t,t-1),t-1,t-2)
\end{equation}
From this definition, we can determine the target $\hat{x}$ that makes the equality hold (see Appendix \ref{appendix:target}):
\begin{equation}
\hat{x} = \frac{x_{t-2}\sqrt{1 - \gamma_t} - x_t \sqrt{1 - \gamma_{t-2}}}{\sqrt{\gamma_{t-2}}\sqrt{1 - \gamma_t} - \sqrt{\gamma_t}\sqrt{1 - \gamma_{t-2}}}\label{eq:x02}
\end{equation}
The signal loss is inferred by rewriting the noise prediction error (see Appendix \ref{appendix:loss}), yielding:
\begin{equation}
\mathcal{L}(\phi)=\frac{\gamma_t}{1-\gamma_t}\left\|g_\phi(x_t,t) - \hat{x}\right\|_2^2
\end{equation}
Once a student has been trained to completion, it becomes the teacher and the process is repeated until the final model has the desired number of steps.
$\log_2T$ training phases are required to distill a $T$-steps teacher to a single-step model and each trained student requires half the sampling steps of its teacher to generate high-quality samples.

%% file: method.tex
\section{Method}
We propose TRAnsitive Closure Time-Distillation (\tcd), an extension of BTD, that reduces the number of distillation phases from $\log_2T$ to a small constant, typically $1$ or $2$.
We focus on the VP setting used in BTD first, but the method itself is independent of it and we illustrate it in the Variance Exploding (VE) setting at the end of the section.
While TRACT also works for noise-predicting objectives, we demonstrate it on signal-prediction where the neural network predicts an estimate of $x_0$.

\subsection{Motivation}
We conjecture that the final quality of samples from a distilled model is influenced by the number of distillation phases and the length of each phase.
As later validated in the experiments section, we consider two potential explanations as to why it is the case.

\subsubsection*{Objective degeneracy} \label{objective_degeneracy}
In BTD, the student in the previous distillation phase becomes the teacher for the next phase.
The student from the previous phase has a positive loss which yields an imperfect teacher for the next phase.
These imperfections accumulate over successive generations of students which leads to objective degeneracy.

\subsubsection*{Generalization} \label{generalization}
Stochastic Weight Averaging (SWA) has been used to improve the performance of neural networks trained for DDPMs~\citep{ddpm}.
With Exponential Moving Average (EMA), the momentum parameter is limited by the training length: high momentum yields high-quality results but leads to over-regularized models if the training length is too short.
This ties in with the time-distillation problem since the total training length is directly proportional to the number of training phases.

\subsection{TRACT}

\begin{figure}[h]
  \centering
  \includegraphics[width=0.8\textwidth]{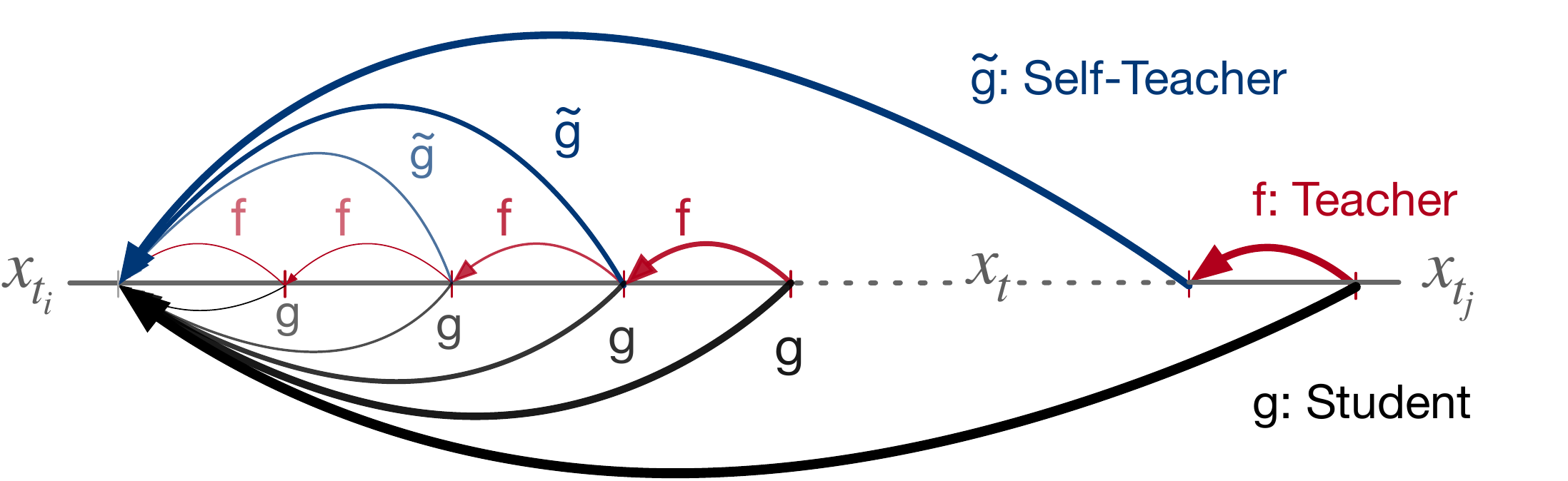}
  \caption{Transitive Closure Distillation of a group $\{t_i,\ldots,t_j\}$.}
  \label{fig:whale}
\end{figure}

\tcd{} is a multi-phase method where each phase distills $T$-steps schedule to $T' < T$ steps, and is repeated until the desired number of steps is reached.
In a phase, the $T$-steps schedule is partitioned into $T'$ contiguous groups.
The partitioning strategy is left open; for example, in our experiments we used equally-sized groups as demonstrated in Algorithm (\ref{alg:TCTD}).

Our method can be seen as an extension of BTD which is not constrained by $T'=T/2$.
However, computational implications arise from the relaxation of this constraint, such as the estimation of $x_{t'}$ from $x_t$ for $t' < t$.

For a contiguous segment $\{t_i,\ldots,t_j\}$, we model the student $g_\phi$ to jump to step $t_i$ from any step $t_i < t \leq t_j$ as illustrated in Figure (\ref{fig:whale}):
\begin{equation}
     \delta(g_\phi,x_t,t,t_i) = \delta(f_\theta, \delta(f_\theta, \ldots\delta(f_\theta,x_t,t,t-1), \ldots), t_{i+1}, t_i)\label{eq:f-f-f}
\end{equation}
The student $g$ is specified to encompass $(t_j-t_i)$ denoising steps of $f$. However, this formulation could require multiple calls of $f$ during training, leading to prohibitive computational costs.

To resolve this issue, we use a self-teacher whose weights are an exponential moving average (EMA) \citep{swa} of the student $g$. This approach is inspired from semi-supervised learning~\citep{lee2013pseudo}, reinforcement learning~\citep{rl_ema} and representation learning~\citep{grill2020bootstrap}. 
For a student network $g$ with weights $\phi$, we denote the EMA of its weights as $\tilde \phi=\texttt{EMA}(\phi,\mu_S)$ where $\mu_S \in [0, 1]$, the momentum, is an hyper-parameter.

The transitive closure operator can now be modeled with self-teaching by rewriting the closure in Equation (\ref{eq:f-f-f}) as a recurrence:
\begin{equation}
     \delta(g_\phi,x_t,t,t_i) \approx x_{t_i} \coloneqq \delta(g_{\tilde\phi}, \delta(f_\theta,x_t,t,t-1), t-1, t_i)
\end{equation}

From this definition, we can determine the target $\hat{x}$ that makes the equality hold using the same method as for Equation (\ref{eq:x02}), see Appendix \ref{appendix:target} for details:
\begin{equation}
\hat{x} = \frac{x_{t_i}\sqrt{1 - \gamma_t} - x_t \sqrt{1 - \gamma_{t_i}}}{\sqrt{\gamma_{t_i}}\sqrt{1 - \gamma_t} - \sqrt{\gamma_t}\sqrt{1 - \gamma_{t_i}}}
\label{eq:xti}
\end{equation}

For the special case $t_i=t-1$, we have $\hat{x}=f_\theta(x_t, t)$.

The loss is the standard signal-predicting DDIM distillation training loss, e.g. for a target value $\hat{x}$:
\begin{equation}
\mathcal{L}(\phi)=\frac{\gamma_t}{1-\gamma_t}\left\|g_\phi(x_t,t)-\hat{x}\right\|_2^2
\end{equation}

\begin{algorithm*}[t]
        \caption{Single-phase \tcd{} training from T timesteps to T/S timesteps for groups of size S.}
   \label{alg:TCTD}
\begin{algorithmic}[1]
\STATE{\textbf{Inputs} Training data $\mathcal{X}$, time schedule $\gamma\in\mathbb{R}^T$, $f_\theta$ is the teacher, $g_\phi$ is the student being trained}
\STATE{$\phi\leftarrow\theta; \; \tilde{\phi}\leftarrow\theta$} \COMMENT{\text{Initialize self-teacher and student from teacher}}
\FOR{batch of training data $\big(x^{(b)}; b \in \{1, \ldots, B\}\big) \sim \mathcal{X}$}
\STATE{$\mathcal{L}(\phi) \leftarrow 0$}
\FOR{$b = 1$ \TO $B$}
    \STATE{\textbf{sample }$\epsilon\sim \mathcal{N}(0,1)$}\COMMENT{Same shape as $x^{(b)}$} 
    \STATE{\textbf{sample }$s\sim \{0,S,2S,\ldots,T-S\}$} \COMMENT{Sample the group starting position}
    \STATE{\textbf{sample }$p\sim \{1,\ldots,S\}$} \COMMENT{Sample the index within the group}
    \STATE{$t\leftarrow s + p$} \COMMENT{Time step to distill}
    \STATE{$x_t\leftarrow x^{(b)}\sqrt{\gamma_t}+\epsilon\sqrt{1-\gamma_t}$} \COMMENT{Generate noisy sample}
    \STATE{$x_{t-1}\leftarrow(x_t \sqrt{1-\gamma_{t-1}} + f_\theta(x_t,t)(\sqrt{\gamma_{t-1}(1-\gamma_t)}-\sqrt{\gamma_t(1-\gamma_{t-1})}))/\sqrt{1-\gamma_t}$}
    \IF{$s = t-1$}
         \STATE{$x_s\leftarrow x_{t-1}$}
    \ELSE
         \STATE{$x_s\leftarrow(x_{t-1} \sqrt{1-\gamma_s} + g_{\tilde\phi}(x_{t-1},t-1)(\sqrt{\gamma_s(1-\gamma_{t-1})}-\sqrt{\gamma_{t-1}(1-\gamma_s)}))/\sqrt{1-\gamma_{t-1}}$}
    \ENDIF
    \STATE{target $\hat{x}\leftarrow(x_s\sqrt{1 - \gamma_t} - x_t \sqrt{1 - \gamma_s})/(\sqrt{\gamma_s}\sqrt{1 - \gamma_t} - \sqrt{\gamma_t}\sqrt{1 - \gamma_s})$}
    \STATE{$\mathcal{L}(\phi) \leftarrow \mathcal{L}(\phi) +  \max(1,\gamma_t/(1-\gamma_t))\frac{1}{B}\left\|g_\phi(x_t,t) - \texttt{stop\_gradient} (\hat{x})\right\|_2^2$} 
\ENDFOR
\STATE{Compute gradients of $\mathcal{L}(\phi)$ and update parameters $\phi$}
\STATE{Update $\tilde\phi\leftarrow EMA(\phi,\mu_S)$}\COMMENT{$\mu_S$ is a hyper-parameter for EMA momentum}
\ENDFOR
\end{algorithmic}
\end{algorithm*}

\subsection{Adapting TRACT to a Runge-Kutta teacher and Variance Exploding noise schedule}

To illustrate its generality, we apply \tcd{} to teachers from Elucidating the Design space of diffusion Models (EDM) \citep{karras2022elucidating} that use a VE noise schedule and an RK sampler.

\paragraph{VE noise schedules} A VE noisification process is parameterized by a sequence of noise standard deviations $\sigma_t \ge 0$ for $t \in \{1,...,T\}$ with $\sigma_1 = \sigma_{min} \le \sigma_t \le \sigma_T = \sigma_{max}$, and $t=0$ denotes the noise-free step $\sigma_0 = 0$. A noisy sample $x_t$ is produced from an original sample $x_0$ and Gaussian noise $\epsilon$ as follows:
\begin{equation} \label{eq:VE}
     x_t = x_0 + \sigma_t \epsilon
\end{equation}
\paragraph{RK step function}
Following on the EDM approach, we use an RK sampler for the teacher and distill it to a DDIM sampler for the student.
The corresponding step functions are $\delta_{RK}$ and $\delta_{DDIM-VE}$, respectively.
The $\delta_{RK}$ step function to estimate $x_{t'}$ from $x_t$, $t > 0$, is defined as:
\begin{equation}
    \delta_{RK}(f_\theta,x_t,t,t') \coloneqq \begin{cases}
    x_t + (\sigma_{t'} - \sigma_{t}) \epsilon(x_t, t) & \text{ if } t' = 0\\
    x_t + \frac{1}{2} (\sigma_{t'} - \sigma_{t}) \left[\epsilon(x_t, t) + \epsilon(x_t + (\sigma_{t'} - \sigma_{t}) \epsilon(x_t, t), t)\right]& \text{otherwise}
    \end{cases}
\end{equation}
where $\epsilon(x_t, t) \coloneqq \frac{x_t - f_\theta(x_t, t)}{\sigma_t}$.

The $\delta_{DDIM-VE}$ step function to estimate $x_{t'}$ from $x_t$, $t > 0$, is defined as:
\begin{equation}
    \delta_{DDIM-VE}(f_\theta,x_t,t,t') \coloneqq f_\theta(x_t,t) \left(1 - \frac{\sigma_{t'}}{\sigma_t}\right) + \frac{\sigma_{t'}}{\sigma_{t}} x_t
\end{equation}
Then, learning the transitive closure operator via self-teaching requires:
\begin{equation}
     \delta_{DDIM-VE}(g_\phi,x_t,t,t_i) \approx x_{t_i} \coloneqq \delta_{DDIM-VE}(g_{\tilde\phi}, \delta_{RK}(f_\theta,x_t,t,t-1), t-1, t_i)
\end{equation} \label{eq:closure_EDM}
From this definition, we can again determine the target $\hat{x}$ that makes the equality hold:
\begin{equation}
\hat{x}  = \frac{\sigma_t x_{t_i} - \sigma_{t_i} x_t}{\sigma_t - \sigma_{t_i}}
\end{equation}
The loss is then a weighted loss between the student network prediction and the target. We follow the weighting and network preconditioning strategies introduced in the EDM paper \citep{karras2022elucidating}:
\begin{equation}
\mathcal{L}(\phi) = \lambda(\sigma_t) \| g_\phi(x_t, t) - \hat{x}\|_2^2
\end{equation}
The resulting distillation algorithm and details on the derivation of $\delta_{RK}$, $\delta_{DDIM-VE}$ as well as the training target $\hat{x}$ can be found in Appendix \ref{appendix:EDM}.

%% file: experiments.tex
\section{Experiments}

We present results with \tcd{} on two image generation benchmarks: CIFAR-10 and class-conditional 64x64 ImageNet.
On each dataset, we measure the performance of our distilled models using the Frechet Inception Distance (FID)\citep{fid}, computed from 50,000 generated samples.
We run each experiment with three seeds to compute the mean and standard deviation.
1-step \tcd{} models improve FID from 9.1 to 4.5 on CIFAR-10 and from 17.5 to 7.4 on 64x64 ImageNet compared to their BTD \citep{binarydistill} counterparts, using the exact same architecture and teacher models. We also present results with \tcd{} when distilling EDM teacher models \citep{karras2022elucidating} using a RK sampler and VE noise schedule: they further improve our FID results to 3.8 on CIFAR-10, see Table (\ref{table:cifar_fid-main}).

We follow up with ablations of the key components of our method: momentums for self-teaching and inference EMAs, and distillation schedules.

\subsection{Image generation results with BTD teachers}

The teacher model in each \tcd{} distillation experiment is initialized from teacher checkpoints of the BTD paper \cite{binarydistill}\footnote{\url{https://github.com/google-research/google-research/tree/master/diffusion_distillation}} so as to be directly comparable to them.

We use a two-phase $T:1024\to32\to1$ distillation schedule.
At the start of each phase, the student's weights are initialized from the current teacher being distilled.
In the first phase, the teacher model uses a 1024-step sampling schedule and the student learns to generate samples in 32 steps.
In the second phase, the teacher is initialized as the student from the previous phase, and the student learns to generate images in a single step.

\paragraph{CIFAR-10} We experimented with two training lengths: 96M samples to match the BTD \citep{binarydistill} paper, and 256M samples to showcase the benefits of longer training with \tcd{}.
Our 1-step \tcd{-96M} model obtains an FID of 5.02 that cuts in almost half the previous state-of-the-art of 9.12 \citep{binarydistill} with the same architecture and training budget.
\tcd{-256M} further improves our 1-step FID results to 4.45. 
For both training budgets, we also run distillation experiments ending with a larger number of steps: $T:1024\to32\to K$ with $K \in \{2, 4, 8\}$ and obtain state-of-the-art models at all steps.
1 and 2 step results are presented on Table \ref{table:cifar_fid-main} while 4 and 8 step results are presented on Table \ref{table:cifar_fid}. More experimental details can be found in Appendix \ref{appendix:tract_details}.

\begin{table}[]
\parbox{.9\linewidth}{
    \centering
    \begin{tabular}{@{}lclc@{}}
    Method     & NFEs & FID  & Parameters \\ \midrule
    TRACT-EDM-256M$^*$  &   1   & \textbf{3.78}  $\pm 0.01$ & 56M \\
    DFNO \citep{zheng2022fast}$^\dag$      &      & 4.12 & 65.8M \\
    TRACT-EDM-96M$^*$  &      & 4.17  $\pm 0.03$ & 56M \\
    TRACT-256M &      & 4.45 $\pm 0.05$ & 60M\\
    TRACT-96M &      & 5.02 $\pm 0.04$ & 60M \\
    BTD-96M \citep{binarydistill}        &     & 9.12  & 60M  \\ \midrule
    TRACT-256M &   2   & \textbf{3.32} $\pm 0.02$ & 60M \\
    TRACT-96M  &      & 3.53 $\pm 0.03$ & 60M \\
    TRACT-EDM-256M$^*$  &      & 3.55  $\pm 0.01$ & 56M \\
    TRACT-EDM-96M$^*$  &      & 3.75  $\pm 0.02$ & 56M\\
    BTD-96M \citep{binarydistill}        &     & 4.51 & 60M    \\
    \end{tabular}
    \caption{FID results on CIFAR-10. $\dag$ Diffusion Fourier Neural Operators (DFNO) use a different model architecture for the student network and require generating a synthetic dataset for training. $*$ \tcd{}-EDM models use better teachers.
    }
    \label{table:cifar_fid-main}
}
\end{table}

\begin{table}[]
\parbox{.9\linewidth}{
    \centering
    \begin{tabular}{@{}lclc@{}}
    Method     & NFEs & FID  & Parameters \\ \midrule
    TRACT-96M &    1  & \textbf{7.43} $\pm$ 0.07 & 296M \\
    TRACT-EDM-96M$^*$  &      & 7.52  $\pm$ 0.05 & 296M \\
    DFNO \citep{zheng2022fast}$^\dag$      &      & 8.35 & 329M \\
    BTD-1.2B \citep{binarydistill}        &     &  17.5 & 296M  \\\midrule
    TRACT-EDM-96M$^*$  &    2  &  \textbf{4.97} $\pm$ 0.03  & 296M\\
    TRACT-96M  &      &  5.24 $\pm$ 0.02  & 296M \\
    BTD-1.2B \citep{binarydistill}        &     & 7.2 & 296M    \\
    \end{tabular}
    \caption{FID results on 64x64 ImageNet. $\dag$ Diffusion Fourier Neural Operators (DFNO) use a different model architecture for the student network and require generating a synthetic dataset for training. $*$ \tcd{}-EDM models use better teachers.}
    \label{table:imagenet_fid-main}
}
\end{table}
\paragraph{64x64 ImageNet} On class-conditional 64x64 ImageNet, our single-step \tcd{}-96M student achieves a FID of 7.43, which improves our BTD counterpart by 2.4x. Due to resource constraints, we did not distill a \tcd{} model with as many training samples (1.2B) as BTD~\citep{binarydistill}. Therefore, the new state-of-the-art that we set on the same model architecture is obtained with a tenth of the training budget. 1 and 2 step results are presented in Table \ref{table:imagenet_fid-main} while 4 and 8 step results are presented on Table \ref{table:imagenet_fid}. More experimental details can be found in Appendix \ref{appendix:tract_details}.
\subsection{Image generation results with EDM teachers}
EDM models \citep{karras2022elucidating} are initialized from checkpoints released with the paper\footnote{\url{https://nvlabs-fi-cdn.nvidia.com/edm/pretrained/}}, which are based off NCSN++ architecture\cite{song_sde} for CIFAR-10, and ADM architecture\cite{dhariwal2021diffusion} for 64x64 ImageNet. 
Results for \tcd{}-EDM models are presented on Table \ref{table:cifar_fid-main} and \ref{table:cifar_fid} for CIFAR-10 as well as Table~\ref{table:imagenet_fid-main} and \ref{table:imagenet_fid} for 64x64 ImageNet. Experimental details can be found in Appendix \ref{appendix:edm_details}.

\subsection{Stochastic Weight Averaging ablations}
\tcd{} uses two different EMAs: one for the self-teacher and one for the student model used at inference time. The self-teacher uses a fast-moving (low momentum) EMA with momentum $\mu_S$ and the inference model uses a slow-moving (high momentum) EMA with momentum $\mu_I$. We study both momentums across ablations on CIFAR-10.

\paragraph{Implementation}
We use a bias-corrected EMA for our experiments.
We detail this implementation for the self-teacher weights $\tilde \phi = \texttt{EMA}(\phi,\mu_S)$.
At the start of training $\tilde \phi_0= \phi_0$, and it is updated at each training step $i > 0$ with:
\begin{equation}
     \tilde \phi_i = \left(1 - \frac{1 - \mu_S}{1 - \mu_S^i}\right) \tilde \phi_{i - 1} + \frac{1 - \mu_S}{1 - \mu_S^i} \phi_i,
\end{equation}
We use the same implementation for the inference model weigths $\texttt{EMA}(\phi, \mu_I)$

\paragraph{Self-teaching EMA}
The momentum parameter $\mu_S$ for the self-teaching EMA strikes a balance between convergence speed and training stability. With low $\mu_S$, the self-teacher weights adapt rapidly to training updates but incorporate noise from the optimization process, leading to unstable self-teaching. On the other hand, higher $\mu_S$ values yield stable self-teaching targets but introduce latency between the student model state and that of its self-teacher. This, in turn, results in outdated self-teacher targets yielding slower convergence.

For the ablation study of $\mu_S$, we fixed the distillation schedule to $T:1024\to32\to1$, the training length to 48M samples per phase and $\mu_I$ to 0.99995. Results are presented in Table \ref{table:sema}\footnote{The best result in the table does not match our best: throughout ablations, for simplicity and at the cost of performance, we did not allocate a larger share of the training budget to the $32\to1$ distillation phase}.
Performance decreases monotonically as the self-teaching EMA grows above a certain threshold (about $0.9$ in this setting), which supports the slower convergence hypothesis for high values of this parameter. Results are equally worse for values at or below 0.01 and present a high variance.
Similarly to observations made in BYOL \citep{grill2020bootstrap}, we found that a wide range of momentum parameter $\mu_S \in [0.1, 0.9]$ values gives good performance. In light of this, we set $\mu_S=0.5$ for all other experiments.

\begin{table}[]
\parbox{.45\linewidth}{
    \centering
    \begin{tabular}{cc}

    Self-teaching EMA & 1 step FID  \\ \midrule
    0.0               & 6.32                                    \\
    0.001             & 6.38                                   \\
    0.01              & 7.29                                    \\
    0.1               & 5.34                                    \\
    0.5               & \textbf{5.24}                            \\
    0.9               & 6.04                                    \\
    0.99              & 7.61                                     \\
    0.999             & 8.30  \\ 
    
    \end{tabular}
    \caption{Self-teaching EMA ablation results on CIFAR-10.}
    \label{table:sema}
}
\parbox{.45\linewidth}{
    \centering
    \begin{tabular}{cc}

    Inference EMA & 1 step FID  \\ \midrule
    0.999          & 6.91                                    \\
    0.9999         & 5.5                                 \\
    0.99995        & \textbf{5.24}                            \\
    0.99999        & 8.73     \\
    \end{tabular}
    \caption{Inference time EMA ablation results on CIFAR-10.}
    \label{table:ema}
}
\end{table}
\paragraph{Inference EMA}
We use a slow-moving EMA of student weights at inference time, which has been shown empirically to yield better test time performance \cite{swa}.
For the ablation study of $\mu_I$, we fix the distillation schedule to $T:1024\to32\to1$, training length per phase to 48M samples and $\mu_S=0.5$.
Results are presented in Table \ref{table:ema}, we observe that values of $\mu_I$ strongly affect performance.
In \ref{appendix:ema_heuristic} we share a heuristic to compute $\mu_I$ values yielding high quality results across experiments and for varying training lengths.

\subsection{Influence of the number of distillation phases}

In the VP setting, we find that \tcd{} performs best when using a 2-phase $T:1024\to32\to1$ distillation schedule. 
Confirming our original conjecture, we observe that schedules with more phases suffer more from \textit{objective degeneracy}.
However,  we observe the worst results were obtained with a single-phase distillation $T:1024\to1$.
In that case, we suspect that due to the long chain of time steps, a phenomenon similar to gradient vanishing is happening.
We present ablation results on CIFAR-10 with distillation schedules of increasing number of phases from 1 to 5: $T:1024\to1,~T:1024\to32\to1,~T:4096\to256\to16\to1,~T:4096\to512\to64\to8\to1,~T:1025\to256\to64\to16\to4\to1$.  

\paragraph{Fixed overall training length} We set $\mu_I=0.99995$, $\mu_S=0.5$ and the overall training length to 96M samples.
Single-step FID results are presented in Table \ref{table:schedules_96M}.
Results clearly get worse with more distillation phases, providing support to the objective degeneracy hypothesis.

\begin{table*}[th]
\centering
\begin{tabular}{cccc}
Distillation schedule                        & Phases                 & Training length           & 1 step FID  \\ \midrule
\multicolumn{1}{c|}{1024, 1}                 & \multicolumn{1}{c|}{1} & \multicolumn{1}{c|}{96M}  &  14.40                                  \\
\multicolumn{1}{c|}{1024, 32, 1}             & \multicolumn{1}{c|}{2} & \multicolumn{1}{c|}{96M}  &  \textbf{5.24}                                   \\
\multicolumn{1}{c|}{4096, 256, 16, 1}        & \multicolumn{1}{c|}{3} & \multicolumn{1}{c|}{96M}  & 6.06                                   \\
\multicolumn{1}{c|}{4096, 512, 64, 8, 1}     & \multicolumn{1}{c|}{4} & \multicolumn{1}{c|}{96M}  & 7.27                                 \\
\multicolumn{1}{c|}{1024, 256, 64, 16, 4, 1} & \multicolumn{1}{c|}{5} & \multicolumn{1}{c|}{96M}  & 8.33                                 \\   
\end{tabular}
\caption{Time Schedule ablations with fixed overall training length on CIFAR-10.}
\label{table:schedules_96M}
\end{table*}

\paragraph{Fixed training length per phase} \tcd{} with 3, 4 and 5 phase distillation schedules is trained again with an increased training budget, now set to 48M samples \textit{per phase}.
1-step FID results are presented in Table \ref{table:schedules_per_phase}.
Many-phase schedules improve their performance but FID scores are still worse than with the 2-phase schedule, despite leveraging the same training budget per distillation phase.
This suggests that the objective degeneracy problem cannot be fully solved at the cost of a reasonably higher training budget. Meanwhile, as seen in previous experiments (see Table (\ref{table:cifar_fid-main})), 2-phase results with 256M samples improved markedly over 96M samples. Therefore, with a fixed training budget, distilling a 2-phase \tcd{} for longer might be the best choice.

\begin{table*}[th]
\centering
\begin{tabular}{cccc}
Distillation schedule                        & Phases                 & Training length           & FID  \\ \midrule 
\multicolumn{1}{c|}{1024, 32, 1}             & \multicolumn{1}{c|}{2} & \multicolumn{1}{c|}{96M}  &  \textbf{5.24}                                   \\
\multicolumn{1}{c|}{4096, 256, 16, 1}        & \multicolumn{1}{c|}{3} & \multicolumn{1}{c|}{144M} & 5.76                                     \\
\multicolumn{1}{c|}{4096, 512, 64, 8, 1}     & \multicolumn{1}{c|}{4} & \multicolumn{1}{c|}{192M} & 6.83                                     \\
\multicolumn{1}{c|}{1024, 256, 64, 16, 4, 1} & \multicolumn{1}{c|}{5} & \multicolumn{1}{c|}{240M} & 7.04             \\
\end{tabular}
\caption{Time Schedule ablations with fixed training length per phase on CIFAR-10.}
\label{table:schedules_per_phase}
\end{table*}

\paragraph{Binary Distillation comparison}
\label{compare-with-btd}
To further confirm that objective degeneracy is the reason why \tcd{} outperforms BTD \citep{binarydistill}, we compare BTD to \tcd{} on the same BTD-compatible schedule: the 10 phases $T:1024\to512\to256\to...\to2\to1$.
We set $\mu_I=0.99995$ and 48M training samples per distillation phase for both experiments.
In this setting, BTD outperforms \tcd{} with an FID of 5.95 versus 6.8. This is additional confirmation that BTD's inferior overall performance may come from its inability to leverage 2-phase distillation schedules.
Besides the schedule, the other difference between the BTD and \tcd{} is the use of self-teaching by \tcd{}. This experiment also suggests that self-teaching may result in less efficient objectives than supervised training.

\subsection{Beyond time distillation}

In addition to reducing quality degradation with fewer sampling steps, \tcd{} can be used for knowledge distillation to other architectures, in particular smaller ones.
Compared to TRACT-96M, we show a degradation from 5.02 to 6.47 FID at 1 sampling step on CIFAR-10 by distilling a model from 60.0M parameters to 19.4M.
For more details, refer to \ref{appendix:size}.

%% file: conclusion.tex
\section{Conclusion}


Generating samples in a single step can greatly improve the tractability of diffusion models.
We introduce TRAnsitive Closure Time-distillation (\tcd{}), a new method that significantly improves the quality of generated samples from a diffusion model in a few steps.
This result is achieved by distilling a model in fewer phases and with stronger stochastic weight averaging than prior methods.
Experimentally, we show that without architecture changes to prior work, \tcd{} improves single-step FID by up to 2.4$\times$.  
Further experiments demonstrate that \tcd{} can also effectively distill to other architectures, in particular to smaller student architectures.
While demonstrated on images datasets, our method is general and makes no particular assumption about the type of data.
It is left to future work to apply it to other types of data.

An interesting extension of \tcd{} could further improve the quality-efficiency trade-off: tpically, distillation steps in DDIMs/DDPMs have maxed out at 8192 due to computational costs of sampling.
Since \tcd{} allows arbitrary reductions in steps between training phases, we could feasibly distill from much higher step counts teachers, where prior methods could not.
This unexplored avenue could open new research into difficult tasks where diffusion models could not previously be applied.

%% file: appendix.tex
\section{Appendix}

\subsection{Deriving the distillation target $\hat{x}$} \label{appendix:target}

We want our student network to match the closure of the teacher steps via self-teaching. If the student network is perfect we have:
\[
\delta(g_\phi,x_t,t,t_i) = \delta(g_{\tilde\phi}, \delta(f_\theta,x_t,t,t-1), t-1, t_i)
\]

We write the right-hand term $x_{t_i}:= \delta(g_{\tilde\phi}, \delta(f_\theta,x_t,t,t-1), t-1, t_i)$ and develop equations for the training target to obtain a perfect student:

\begin{align*}
     x_{t_i} &= \delta(g_\phi,x_t,t,t_i)\\
    x_{t_i}&=\sqrt \gamma_{t_i} g_\phi(x_t, t) + \sqrt{1 - \gamma_{t_i}} \left( \frac{x_t - \sqrt \gamma_t g_\phi(x_t, t)}{\sqrt{1 - \gamma_t}} \right)\\
    x_{t_i}&= \left(\sqrt \gamma_{t_i} - \frac{\sqrt \gamma_t \sqrt{1 - \gamma_{t_i}}}{\sqrt{1 - \gamma_t}} \right)g_\phi(x_t, t) + \frac{\sqrt{1 - \gamma_{t_i}}}{\sqrt{1 - \gamma_t}} x_t\\
    x_{t_i} \sqrt{1 - \gamma_t}&= \left(\sqrt \gamma_{t_i} \sqrt{1 - \gamma_t} - \sqrt \gamma_t \sqrt{1 - \gamma_{t_i}}\right)g_\phi(x_t, t) + \sqrt{1 - \gamma_{t_i}} x_t\\
    g_\phi(x_t, t) &= \frac{x_{t_i} \sqrt{1 - \gamma_t} - x_t \sqrt{1 - \gamma_{t_i}}}{\sqrt \gamma_{t_i} \sqrt{1 - \gamma_t} - \sqrt \gamma_t \sqrt{1 - \gamma_{t_i}}}
\end{align*}

\subsection{Deriving the distillation loss} \label{appendix:loss}

The signal training loss is derived from the noise training loss between the expected noise $\epsilon$ and predicted noise $\epsilon_{|t}$ as follows:
\begin{align*}
\mathcal{L}(\phi) &= \left\|\epsilon_{|t} - \epsilon\right\|_2^2 \\
 &= \left\|\cfrac{x_t - g_{\phi}(x_t,t) \sqrt{\gamma_t}}{\sqrt{1-\gamma_t}} - \cfrac{x_t - \hat{x}\sqrt{\gamma_t}}{\sqrt{1-\gamma_t}}\right\|_2^2 \\
 &= \cfrac{\gamma_t}{1-\gamma_t}\left\|g_{\phi}(x_t,t) - \hat{x}\right\|_2^2 \\
\end{align*}

In code implementation, it is common for numerical reasons to use:
\begin{align*}
\mathcal{L}(\phi)=\max\left(1,\cfrac{\gamma_t}{1-\gamma_t}\right)\left\|g_{\phi}(x_t,t) - \hat{x}\right\|_2^2
\end{align*}

\subsection{Experimental details for \tcd{}}
\label{appendix:tract_details}
\paragraph{CIFAR-10} To obtain our best performing \tcd{-96M} and \tcd{-256M}, we use a global batch size of 256 split across 8 GPUs and the Adam optimizer with a constant learning rate of $2\times 10^{-4}$, no weight decay, no dropout, and gradient clipping to a norm of 1.0.

The momentum parameter of the self-teaching EMA $\tilde \phi$ is set to $\mu_S=0.5$ during distillation. At sampling time, we use another EMA of the student weights to generate images, whose momentum parameter $\mu_I$ is set to a higher value: $0.99997$ for \tcd{-96M} and $0.99999$ for \tcd{-256M}. Finally, like in \citep{binarydistill} we allocate higher distillation budgets to the $32 \to 1$ than to the $1024 \to 32$ phase. We suspect that the optimal balance between the two stages depends on the overall training length, but did not overly tune this hyperparameter and split total training samples $1/6$ for phase $1024\to32$ and $5/6$ for phase $32\to1$ in both experiments.

\paragraph{64x64 ImageNet} We use a global batch size of 256 split across 8 GPUs for our best performing \tcd{-96M}. We use Adam optimizer with a constant learning rate of $2\times 10^{-4}$, no weight decay, no dropout, and gradient clipping to a norm of 1.0. 

The momentum parameter of the self-teaching EMA $\tilde \phi$ is set to $\mu_S=0.5$ during distillation. At sampling time, we use another EMA of the student weights to generate images, whose momentum parameter $\mu_I$ is set to a higher value of $0.99995$ for \tcd{-96M}. We evenly split the overall training length between phase $1024\to32$ and phase $32\to1$.

Unlike for CIFAR10 experiments, we predict both signal $x_0$ and noise $\epsilon_0$ during distillation, following the setup of BTD~\cite{binarydistill}.

\subsection{Experimental details for \tcd{} with EDM teachers} \label{appendix:edm_details}

\paragraph{CIFAR-10} We use NCSN++ by Song et. al. \cite{song_sde} and pretrained weights from Karras et. al. \cite{karras2022elucidating}. Without augmentation, our model contains 56 million trainable parameters. We use a global batch size of 512 split across 8 GPUs for our best performing \tcd{-96M} and \tcd{-256M}. We use Adam optimizer with a constant learning rate of $1\times10^{-3}$ for one-step distillation, and learning rate of $3\times10^{-4}$ for multi-step distillation. For all settings, we disable weight decay, learning rate warmup, dropout, augmentation, and gradient clipping.

As EDM samplers require lower NFEs compared to typical DDIM samplers, we tune the hyperparameter of timesteps to distill from. We find that distilling from 40 steps (79 NFEs with Runge-Kutta) to 1 step gives a good balance between generating high quality targets from teachers, and ease of learning for the student.

\paragraph{64x64 ImageNet} For class-conditional 64x64 ImageNet, we use ADM model by Dhariwal and Nichol \cite{dhariwal2021diffusion} with 296 million parameters and weights from Karras et. al. \cite{karras2022elucidating} with no changes. We train the model with a global batch size of 512 split across 8 A100 GPUs, with Adam optimizer of a constant learning rate of $3\times10^{-4}$ with a linear learning rate warm-up of 4M samples. We disable weight decay, dropout, augmentation, and gradient clipping as we do not observe impacts on the final FID scores. As 64x64 ImageNet uses a different model targeted for higher resolutions, we apply a one-phase distillation from 128 timesteps (255 NFEs with Runge-Kutta) to one-step or multi-step for improved targets from the teacher.

\subsection{More experimental results}
In Table~\ref{table:cifar_fid} and~\ref{table:imagenet_fid}, we show the results of distilling to 4-step and 8-step student models on CIFAR10 and 64x64 ImageNet.
Performance is slightly worse than BTD~\cite{binarydistill}, possibly due to self-teaching being less efficient than supervised training as discussed in Section~\ref{compare-with-btd}.
\begin{table}[]
\parbox{.9\linewidth}{
    \centering
    \begin{tabular}{@{}lclc@{}}
    Method     & NFEs & FID  & Parameters \\ \midrule
    TRACT-256M &   4   & \textbf{2.93} $\pm 0.04$ & 60M\\
    BTD-96M \citep{binarydistill}         &     & 3.00   & 60M \\
    TRACT-96M  &      & 3.16 $\pm 0.14$ & 60M\\
    TRACT-EDM-256M$^*$  &      & 3.21  $\pm 0.01$ & 56M\\
    TRACT-EDM-96M$^*$  &      & 3.41  $\pm 0.02$ & 56M \\ \midrule
    TRACT-EDM-256M$^*$  &  8    & \textbf{2.41} $\pm 0.01$ & 56M  \\
    TRACT-EDM-96M$^*$  &      & 2.48  $\pm 0.01$ & 56M\\
    BTD-96M \citep{binarydistill}         &     & 2.57   & 60M \\
    TRACT-96M  &      & 2.80 $\pm 0.05$ & 60M\\
    TRACT-256M &      & 2.84 $\pm 0.01$ & 60M\\
    \end{tabular}
    \caption{FID results on CIFAR-10. $\dag$ Diffusion Fourier Neural Operators use a different model architecture for the student network and require generating a synthetic dataset for training. $*$ \tcd{}-EDM models use better teachers.}
    \label{table:cifar_fid}
}
\end{table}

\begin{table}[]
\parbox{.9\linewidth}{
    \centering
    \begin{tabular}{@{}lclc@{}}
    Method     & NFEs & FID  & Parameters \\ \midrule
    BTD-1.2B \citep{binarydistill}         & 4    &  \textbf{3.9} & 296M \\
    TRACT-96M  &      & 4.32 $\pm$ 0.07 & 296M\\
    TRACT-EDM-96M$^*$  &      & 5.25  $\pm 0.32 $ & 296M \\\midrule
    BTD-1.2B \citep{binarydistill}         & 8    &  \textbf{2.9}  & 296M \\
    TRACT-96M  &      & 3.88 $\pm$ 0.05  & 296M\\
    TRACT-EDM-96M$^*$  &      & 3.93  $\pm$ 0.06 & 296M\\
    \end{tabular}
    \caption{FID results on 64x64 ImageNet. $\dag$ Diffusion Fourier Neural Operators use a different model architecture for the student network and require generating a synthetic dataset for training. $*$ \tcd{}-EDM models use better teachers.}
    \label{table:imagenet_fid}
}
\end{table}
\subsection{Generated samples}

\begin{figure}%
    \centering
    \subfloat[\centering 1 sampling step]{{\includegraphics[width=3cm]{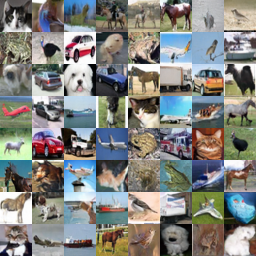} }}%
    \quad
    \subfloat[\centering 2 sampling steps]{{\includegraphics[width=3cm]{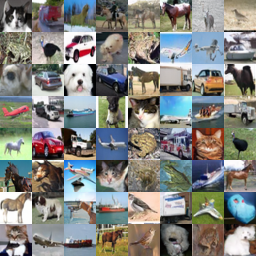} }}%
    \quad
    \subfloat[\centering 4 sampling steps]{{\includegraphics[width=3cm]{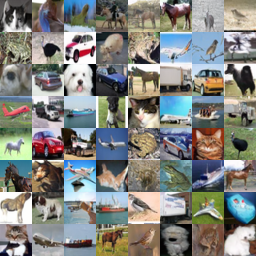} }}%
    \quad
    \subfloat[\centering 8 sampling steps]{{\includegraphics[width=3cm]{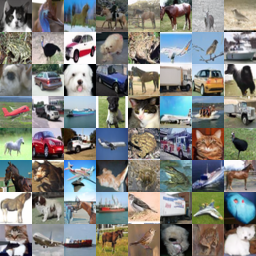} }}%
    \caption{Random samples from \tcd{}-256M on CIFAR10, for a fixed initial noise with varying number of sampling steps.}%
    \label{fig:samples-cifar}%
\end{figure}

\begin{figure}%
    \centering
    \subfloat[\centering 1 sampling step]{{\includegraphics[width=3cm]{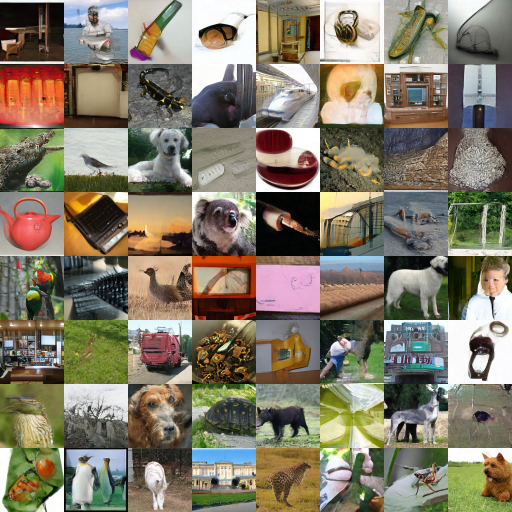} }}%
    \quad
    \subfloat[\centering 2 sampling steps]{{\includegraphics[width=3cm]{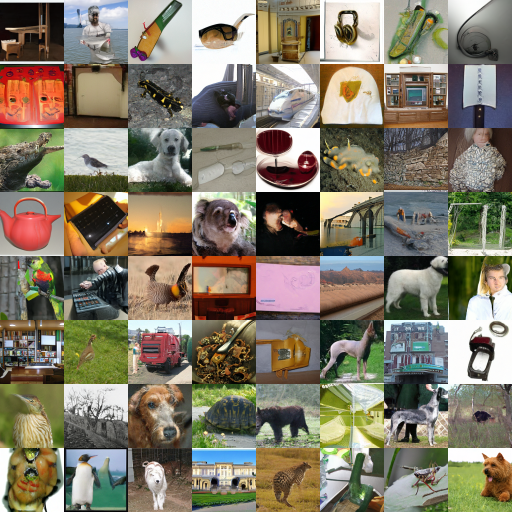} }}%
    \quad
    \subfloat[\centering 4 sampling steps]{{\includegraphics[width=3cm]{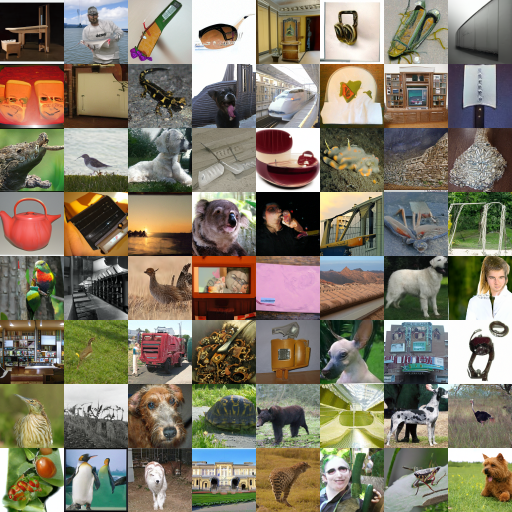} }}%
    \quad
    \subfloat[\centering 8 sampling steps]{{\includegraphics[width=3cm]{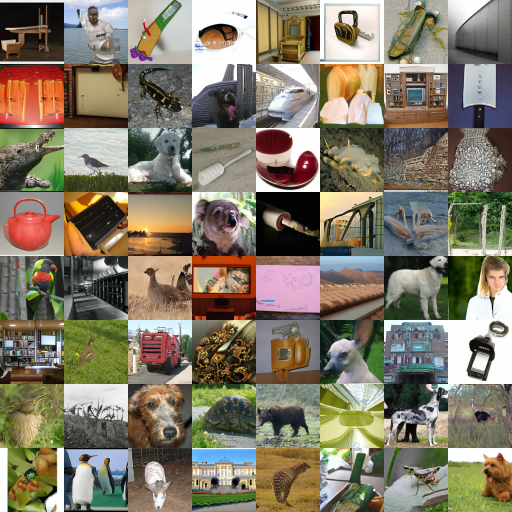} }}%
    \caption{Random samples from \tcd{}-96M on 64 × 64 ImageNet, for a fixed initial noise and class label with varying number of sampling steps.}%
    \label{fig:samples-imagenet-random-class}%
\end{figure}

\begin{figure}%
    \centering
    \subfloat[\centering 1 sampling step]{{\includegraphics[width=3cm]{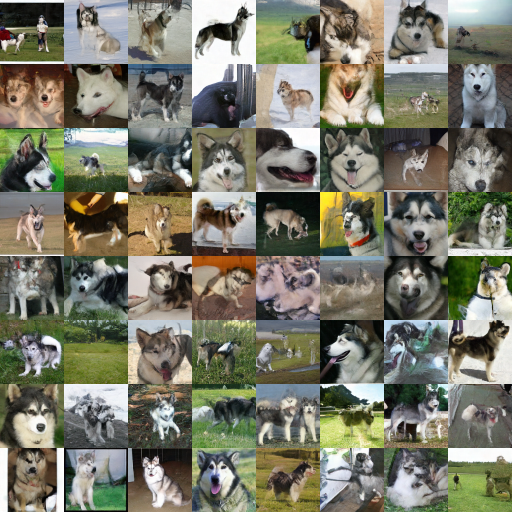} }}%
    \quad
    \subfloat[\centering 2 sampling steps]{{\includegraphics[width=3cm]{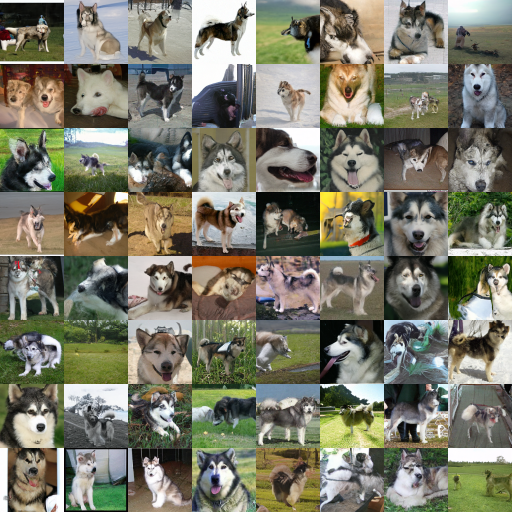} }}%
    \quad
    \subfloat[\centering 4 sampling steps]{{\includegraphics[width=3cm]{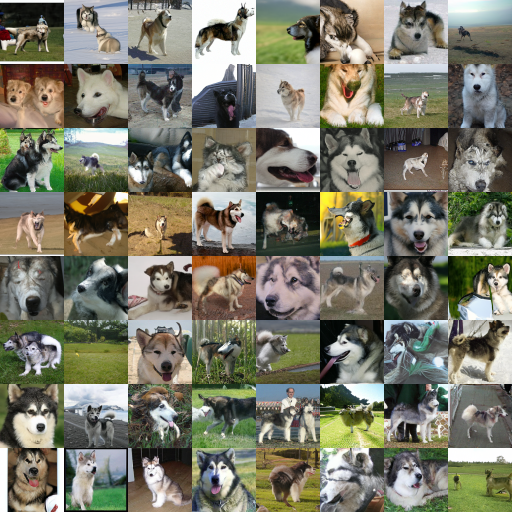} }}%
    \quad
    \subfloat[\centering 8 sampling steps]{{\includegraphics[width=3cm]{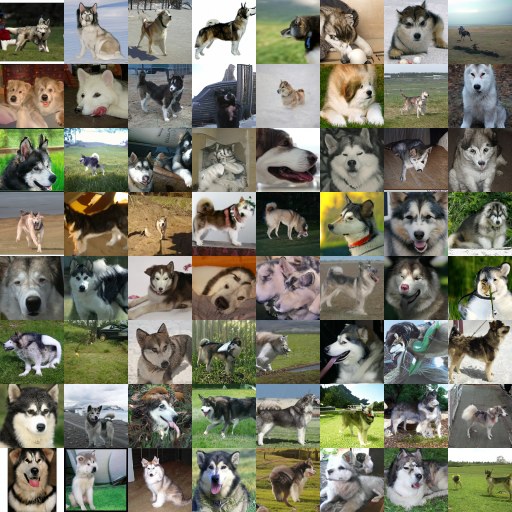} }}%
    \caption{Random samples from \tcd{}-96M on 64 × 64 ImageNet, conditioned on the ‘malamute’ class, for a fixed initial noise with varying number of sampling steps.}%
    \label{fig:samples-imagenet-fixed-class}%
\end{figure}

We present random samples from our distilled models with varying sampling steps.
As shown in Figure~\ref{fig:samples-cifar}, \ref{fig:samples-imagenet-random-class} and \ref{fig:samples-imagenet-fixed-class}, the deterministic mapping between noises and samples are mostly preserved in these distilled models. We can see that there is a slight degrade of image quality for the 1-step student compared with students distilled to more sampling steps.

\subsection{A heuristic to pick the EMA momentum parameter} \label{appendix:ema_heuristic}

In our experiments, we evidenced that $\mu_I$, the EMA momentum parameter for our evaluation model, is key to good performance and must be tuned carefully. This motivated us to come up with a heuristic to pick it efficiently.

We hypothesize that this momentum parameter should be as high as possible while keeping the EMA unbiased to model parameter values at initialization: at the end of training, the weight of initial student parameters should be small in the EMA of model parameters. Formally, training a model over $N$ steps we obtain a sequence of model weights $(\theta_i)_{0 \le i \le N}$, where $\theta_0$ represents the model weights at initialization. At the end of training, the resulting EMA of model weights $\theta_{EMA}$ used for inference is obtained from:
\[
\theta_{EMA} = (1 - \mu_I) \sum_{1 \le i \le N} \mu^{N - i} \theta_i + \mu_I^N \theta_0
\]
In that context, we want $\mu_I^N$ to be small at the end of training and we call $\epsilon$ this small value:
\[
\mu_I^N := \epsilon
\]
We parameterize our heuristic this way: when varying our training length, $N$ changes but we keep $\epsilon$ fixed to derive an appropriate $\mu_I$ value used by the inference time EMA. We picked $\epsilon = 10^{-4}$ in our experiments and obtained good results with it.

\begin{figure}[]
  \centering
  \includegraphics[width=0.9\textwidth,height=0.33\textheight]{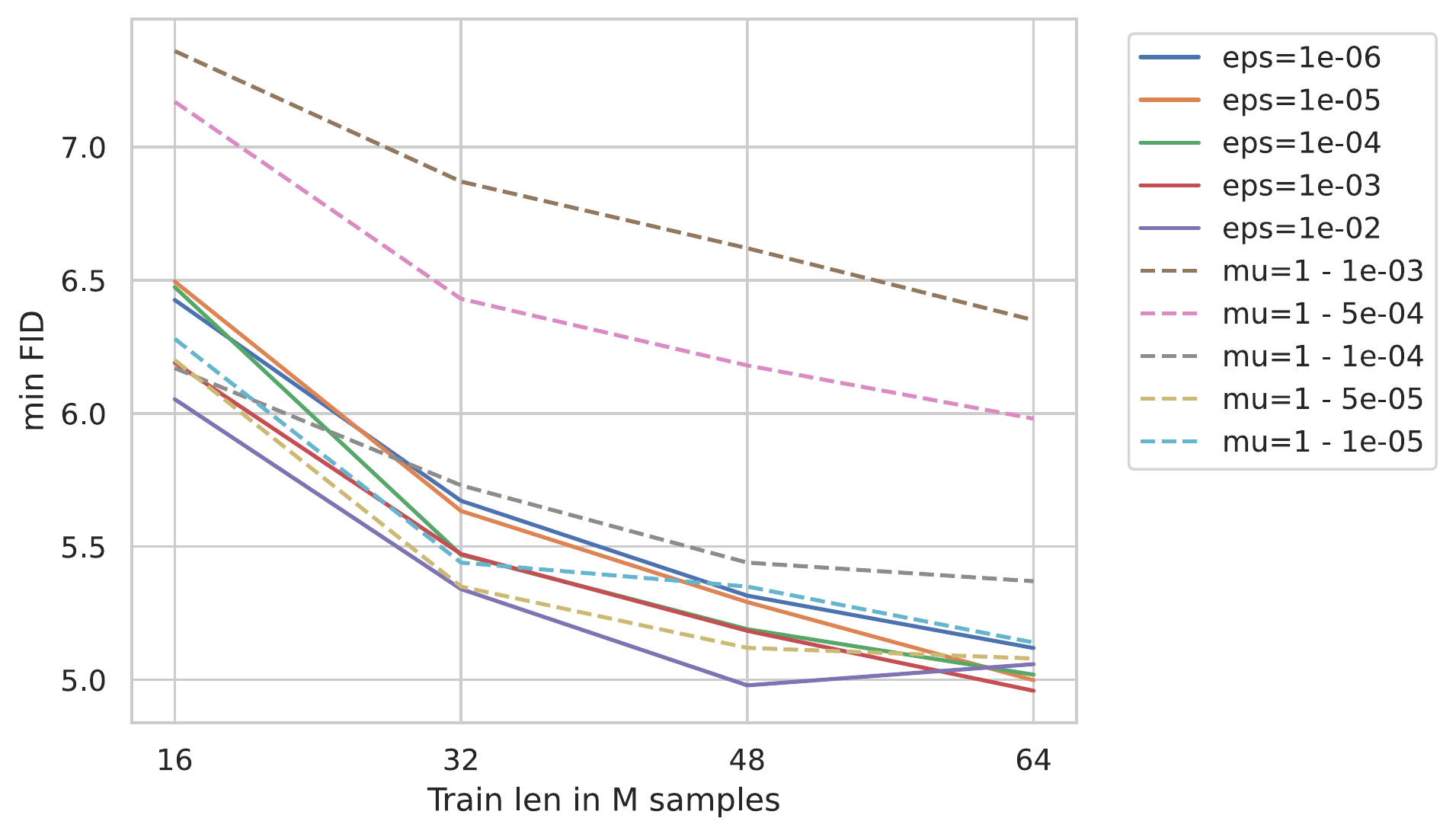}
  \caption{1-step FID for 2-phases $T:1024\to32\to1$ \tcd{} distilled models. Each curve maps to a different way to set the inference time EMA momentum $\mu$ across training lengths. Dashed lines correspond to fixing a $\mu$ value, solid lines correspond to fixing $\epsilon = \mu^N$.}
  \label{fig:ema_heuristic}
\end{figure}

We compare this heuristic to a direct grid search for a fixed $\mu_I$ parameter.
We carry out experiments on CIFAR-10 with 5 different $\epsilon$ values, $\epsilon \in \{10^{-1},~10^{-2},~10^{-3},~10^{-4},~10^{-5}\}$.
For each value, we train 1-step $T:1024\to32\to1$ \tcd{} models over varying training lengths.
We vary the number of samples $S \in \{16M,~32M,~48M,~64M\}$ with a fixed batch size of $256$, the number of training steps $N$ therefore varies as $S/256$.
We compare results to a grid search for $\mu_I$ with 5 different values, $\mu_I \in \{1-10^{-3},~1-5 \times 10^{-4},~1-10^{-4},~1-5 \times 10^{-5},~1-10^{-5}\}$.
Those values are also fixed across training lengths.
Results are summarized in Figure \ref{fig:ema_heuristic}.
First, we note that training for longer always improves performance. Then, we notice that the $\epsilon$ heuristic seems more effective at finding a good performing EMA momentum parameter $\mu_I$ than a direct grid search.
All $\epsilon$ values reach a reasonable performance and the best performance for a given training length is often for one value of $\epsilon$, rather than a $\mu_I$ found from the direct grid search.

\subsection{\tcd{} with EDM teachers}\label{appendix:EDM}

In this section, we expand derivations of the \tcd{} distillation algorithm with EDM \citep{karras2022elucidating} teacher models. We keep their VE noise schedule, an RK sampler with the teacher model, and distill them into a DDIM student network.

\paragraph{DDIM step with VE noise schedule} With a VE noise schedule, starting from a noisy image $x_t,~t> 0$ with a signal predicting network $f_\theta$ we obtain estimates for the signal and the noise as:
$$
     x_{0|t} \coloneqq f_\theta(x_t,t)\textrm{ and }\epsilon_{|t}=\frac{x_t - x_{0|t}}{\sigma_t}
$$
The corresponding DDIM step from $t$ to any $t'$ is therefore:
\begin{align*}
    x_{t'} &= x_{0|t} + \sigma_{t'} \epsilon_{|t} \\
    x_{t'} &= f_\theta(x_t,t) + \sigma_{t'} \frac{x_t - x_{0|t}}{\sigma_t} \\
    \delta_{DDIM-VE}(f_\theta,x_t,t,t') = x_{t'} &= f_\theta(x_t,t) \left(1 - \frac{\sigma_{t'}}{\sigma_t}\right) + \frac{\sigma_{t'}}{\sigma_{t}} x_t
\end{align*}

\paragraph{RK sampling} The ODE defined by the VE noise process \ref{eq:VE} can be solved by numerical integration from step $t$ to step $t' < t$ with a RK solver following the below steps, which define $\delta_{RK}(f_\theta,x_t,t,t')$. 
\begin{equation*}
\begin{split}
    \epsilon_{|t} &= \frac{x_t - f_\theta(x_t, t)}{\sigma_t} \\
    x_{t'} &= x_t + (\sigma_{t'} - \sigma_{t}) \epsilon_{|t} \\
    \epsilon_{|t}^{'} &= \frac{x_{t'} - f_\theta(x_{t'}, t')}{\sigma_{t'}} \\
    x_{t'} &= x_t + (\sigma_{t'} - \sigma_{t}) \frac{\epsilon_{|t} + \epsilon_{|t}^{'}}{2}
\end{split}
\end{equation*}
Where the last two steps are skipped when $t' = 0$.
    
\paragraph{Deriving the distillation target $\hat{x}$} We want our student network to match the closure of the teacher steps via self-teaching. If the student network is perfect we have:
$$
     \delta_{DDIM-VE}(g_\phi,x_t,t,t_i) = \delta_{DDIM-VE}(g_{\tilde\phi}, \delta_{RK}(f_\theta,x_t,t,t-1), t-1, t_i)
$$

We write the right-hand term $x_{t_i}:= \delta_{DDIM-VE}(g_{\tilde\phi}, \delta_{RK}(f_\theta,x_t,t,t-1), t-1, t_i)$ and develop equations for the training target to obtain a perfect student:
\begin{align*}
     x_{t_i} &= \delta_{DDIM-VE}(g_\phi,x_t,t,t_i)\\
    x_{t_i}&=g_\phi(x_t,t) \left(1 - \frac{\sigma_{t_i}}{\sigma_t}\right) + \frac{\sigma_{t_i}}{\sigma_{t}} x_t\\
    x_{t_i} \sigma_{t} &= g_\phi(x_t,t) \left(\sigma_{t} - \sigma_{t_i}\right) + \sigma_{t_i}x_t\\
    g_\phi(x_t,t) &= \frac{x_{t_i} \sigma_{t} - \sigma_{t_i}x_t}{\sigma_{t} - \sigma_{t_i}}
\end{align*}
We now have all the ingredients to write the \tcd{} distillation algorithm with EDM teachers, which is presented in Algorithm \ref{alg:TCEDM one-step train}.

\begin{algorithm*}[t]
        \caption{Single-phase TRACT (VE-EDM) distillation from T timesteps to T/S timesteps for groups of size S.}
   \label{alg:TCEDM one-step train}
\begin{algorithmic}[1]

\STATE{\textbf{Inputs} Training data $\mathcal{X}$, time schedule $\gamma\in\mathbb{R}^T$, $f_\theta$ is the teacher, $g_\phi$ is the student being trained}
\STATE{$\phi\leftarrow\theta; \; \tilde{\phi}\leftarrow\theta$} \COMMENT{\text{Initialize self-teacher and student from teacher}}
\FOR{batch of training data $\big(x^{(b)}; b \in \{1, \ldots, B\}\big) \sim \mathcal{X}$}
\STATE{$\mathcal{L}(\phi) \leftarrow 0$}
\FOR{$b = 1$ \TO $B$}
    \STATE{\textbf{sample} $\epsilon \sim \mathcal{N}(0,1)$}
    \STATE{\textbf{sample} $s \sim \{0, S, 2S, \dots, T-S\})$}  \COMMENT{Sample the group starting position}
    \STATE{\textbf{sample} $p \sim \{1, \dots, S\})$} \COMMENT{Sample the index within the group}
    \STATE{$t \leftarrow s + p$} \COMMENT{Timestep to distill}
    \STATE{$x_t \leftarrow x^{(b)} + \sigma_t \epsilon$} \COMMENT{Generate noisy sample}
    \STATE{\textcolor{blue}{\texttt{\# Step 1. Runge-Kutta Target from the teacher}}}
    \STATE{$\epsilon_{|t} \leftarrow (x_t - f_\theta(x_t,t)) / \sigma_t$}
    \STATE{$x_{t-1} \leftarrow x_t + (\sigma_{t-1} - \sigma_t) \epsilon_{|t}$}
    \IF{$\sigma_{t-1} \neq 0$}
        \STATE{$\epsilon'_{|t} \leftarrow (x_{t-1} - f_\theta(x_{t-1}, t-1)) / \sigma_{t-1}$} \COMMENT{RK second-order correction}
        \STATE{$x_{t-1} \leftarrow x_t + \frac{1}{2} (\sigma_{t-1} - \sigma_t) (\epsilon_{|t} + \epsilon'_{|t})$}
    \ENDIF
    \STATE{\textcolor{blue}{\texttt{\# Step 2. Self-teaching step}}}
    \IF{$s = t - 1$}
        \STATE{$x_s \leftarrow x_{t-1}$}
    \ELSE 
        \STATE{$x_s \leftarrow g_{\tilde \phi}(x_{t-1},t-1) (1 - \sigma_{s}/\sigma_{t-1}) + x_{t-1}\sigma_{s}/\sigma_{t-1}$} \COMMENT{VE DDIM step}
    \ENDIF

    \STATE{\textcolor{blue}{\texttt{\# Step 3. Base case of terminal noise indices}}}
    \IF{$t-1 = 0$}
        \STATE{target $\hat{x} \leftarrow f_\theta(x_t,t)$}
    \ELSE 
        \STATE{target $\hat{x} \leftarrow (\sigma_t x_s - x_t \sigma_s) / (\sigma_t - \sigma_s)$}
    \ENDIF
    \STATE{Loss $\mathcal{L}(\phi) \leftarrow \mathcal{L}(\phi) + \frac{1}{B} \lambda(\sigma_t) \| g_\phi(x_t ,t) -  \texttt{stop\_gradient}(\hat{x}) \|_2^2$} \COMMENT{EDM loss}
    \ENDFOR
    \STATE{Compute gradients of $\mathcal{L}(\phi)$ and update parameters $\phi$}
    \STATE{Update $\tilde\phi \leftarrow EMA(\phi,\mu_S)$}
\ENDFOR
\end{algorithmic}
\end{algorithm*}

\subsection{Knowledge Distillation}
\label{appendix:size}

\tcd{} can also be used for knowledge distillation from architecture A to another architecture B using only one additional training phase.
Of particular interest is the case when the target architecture B has lower computational complexity than A.
The additional training phase is a standard distillation (i.e. not time-based), where the number of steps is held constant while the teacher and student have different architectures.
For instance, a distillation schedule $T:1024\to32\to1$ leads to two possibilities $T:(1024_A\to1024_B)\to32_B\to1_B$ and $T:1024_A\to(32_A\to32_B)\to1_B$. In our experiments, we created the architecture B with 67\% fewer parameters by keeping the architecture A (from BTD \citep{binarydistill} paper), and reducing the number of channels for each layer from 256 to 128. We show comparable performance in Table \ref{table:size} on CIFAR-10 with 1 sampling step.

\begin{table*}[th]
\centering
\begin{tabular}{ ccccc }
Method        & Distillation schedule & Parameters & Training Length & FID                         \\ \hline
BTD           & 1024, 512, ... , 1    & 60.0M      & 96M             & 9.12                        \\
\tcd{} & 1024, 32, 1           & 60.0M      & 96M            & 5.02                        \\
\tcd{} & 1024, 1024, 32, 1     & 19.4M      & 432M            & 7.17                        \\
\tcd{} & 1024, 32, 32, 1       & 19.4M      & 336M            & 6.47
\end{tabular}
\caption{Single step FID on CIFAR-10.}
\label{table:size}
\end{table*}

It could be interesting to validate in future work whether \tcd{} works with heterogeneous types of model architectures for students and teachers, such as Transformers \citep{peebles2022scalable}.